\documentclass[twocolumn]{webofc}
\usepackage[varg]{txfonts}   
\usepackage[square,numbers]{natbib}
\graphicspath{{graphics/}{graphics/arch/}{Graphics/}{./}} 
\begin{document}
\title{An Investigation of the Factors Influencing Evolutionary Dynamics in the Joint Evolution of Robot Body and Control}
%
%

\author{\firstname{Léni K.} \lastname{Le Goff}\inst{1}\fnsep\thanks{\email{l.legoff2@napier.ac.uk}} \and
        \firstname{Edgar} \lastname{Buchanan}\inst{2}\fnsep\thanks{\email{edgar.buchanan@york.ac.uk}} \and
        \firstname{Emma} \lastname{Hart}\inst{1}\fnsep\thanks{\email{e.hart@napier.ac.uk}}
}

\institute{Edinburgh Napier University
\and
           University of York
          }

\abstract{%
In evolutionary robotics, jointly optimising the design and the controller of robots is a challenging task due to the huge complexity of the solution space formed by the possible combinations of body and controller. We focus on the evolution of robots that can be physically created rather than just simulated, in a rich morphological space that includes a voxel-based chassis, wheels, legs and sensors.  On the one hand, this space offers a high degree of liberty in the range of robots that can be produced,  while on the other hand introduces a complexity rarely dealt with in previous works relating to matching controllers to designs and in evolving closed-loop control.  
This is usually addressed by augmenting evolution with a learning algorithm to refine controllers. Although several frameworks exist, few have studied the role of the \textit{evolutionary dynamics} of the intertwined `evolution+learning' processes in realising high-performing robots.  We conduct an in-depth study of the factors that influence these dynamics, specifically: synchronous vs asynchronous evolution; the mechanism for replacing parents with offspring, and rewarding goal-based fitness vs novelty via selection. Results show that asynchronicity  combined with goal-based selection and a `replace  worst' strategy results in the highest performance.}
\maketitle
\begin{figure}[h]
    \centering
    \includegraphics[width=0.5\textwidth]{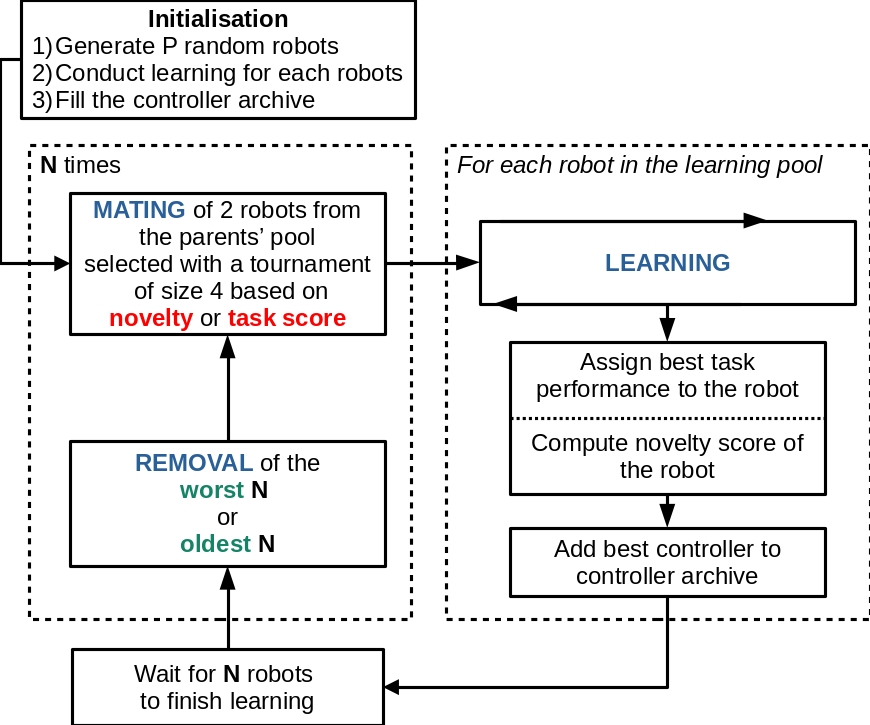}
    \caption{Diagram describing the  evolutionary algorithm studied in this paper. The algorithm is described in details in section~\ref{sec:ea}}
    \label{fig:gen_dia}
\end{figure}
\section{Introduction}\label{sec:intro}

Ever since the pioneering work of Sims in 1994 \cite{sims1994evolving}, evolutionary algorithms have been applied to the task of co-optimising the design and control of robots, i.e simultaneous evolution of both body and brain \cite{hart2022artificial}. The field has grown rapidly with many examples of robots evolved in diverse morphological spaces to complete a broad range of tasks. A recent focus has been directed towards evolving robots in simulation that can be physically built following evolution  (rather than existing only in a digital world). This has led to evolved designs  built for example from living stem-cells \cite{Kriegman1853}, soft-materials \cite{lipson2000automatic} and fixed sets of rigid plastic components \cite{moreno2021emerge}. Most of these examples have open-loop controllers, i.e. they are able to move in a various ways but this is not guided by information sensed directly from the environment. Recent work has tried to address this  (\cite{hale2019robot, le2022morpho}) but demonstrates that evolving a design that operates via a closed loop controller that exploits sensory information is very challenging.



As a result, research in this direction usually combines evolution with some form of learning \cite{EibenHart}, inspired  on the one hand by the observation that the interlinked processes of learning and evolution have resulted in a remarkable diversity of morphological forms within environmental niches within the natural world,  and on the other by the more pragmatic observation that adding a learning process enables an inherited controller to be rapidly tuned to a new body.  The latter point is particularly relevant if evolving closed-loop controllers where there needs to be tight integration between sensors and actuators. It should also be noted that researchers in this field tend to have the dual goal of using the `evolving+learning' framework to evolve robots that are both high-performing \textit{and} diverse with respect to their morphology: the latter is important for several reasons: (1) morpho-evolutionary algorithms tend to converge very quickly to a single solution which maybe sub-optimal \cite{cheney2016difficulty,cheney2018scalable}; (2) evolving diverse designs increases the chance of crossing the reality-gap when transferring from simulation to reality \cite{koos2012transferability} and (3) offers the possibility of escaping from local optima in the behavioural space \cite{buchanan2020bootstrapping}.

A number of frameworks exist for integrating and learning, using a range of architectures. For example, in \cite{miras2020evolving,le2022morpho}, an approach is utilised in which evolution is applied to a population of body designs, with a learning algorithm applied to each new design to determine fitness\footnote{where the learning algorithm may itself be population based, e.g. using CMA-ES to evolve weights of neural controllers}. Once all learning is complete, the entire population is replaced, either selecting from (parents+offspring) or offspring only. Evolution is thus \textit{synchronous}. In contrast, Gupta {\em et al} \cite{gupta2021embodied} suggest a fully 
\textit{asynchronous} architecture that uses a \textit{steady-state} algorithm to make the process more tractable. 
This necessitates definition of a replacement mechanism to determine which member of the population is removed each time a new child becomes available, with the authors opting for removal of the oldest member, arguing that this maintains diversity.  However, to the  best of our knowledge, there has never been a systematic study of how the dynamics of an `evolution+learning' framework is influenced by the choice of architecture, and the impact this has on the final performance and diversity of robots achieved by  the system. 

The contribution of this paper is conduct a study of an `evolution+learning' framework that investigates the influence of three factors on the evolutiony dynamics that determine  the performance and diversity of evolved robots. Specifically we investigate: (1) the method of updating the population as offspring are generated (synchronous or asynchronous); (2) the replacement mechanism (worst or oldest); (3) the reward mechanism (goal-based or novelty based). We include the latter given that searching for novelty has been shown to increase exploration leading to better performance.  All experiments are conducted in a rich morphological space that permits evolution of a free-form chassis, a range of components types (sensors, legs, and wheels) and can result in closed loop control. We use a task in which a robot is evolved to maximise exploration of an environment.

We find that the combination of (asynchronous, replace-worst, goal-based)  evolution leads to the highest performing robots 
In all variants, we find that replacing the worst member(s) of the population is better than replacing using an age-related strategy, in contrast to the argument outlined in \cite{gupta2021embodied}. Our results provide new insights into how these factors influence the dynamics of (evolution+learning) hybrid architectures that can be drawn upon to develop more efficient and effective methods in the future.












%

\section{Related Work}
We discuss previous work in the context of approaches that have considered  joint optimisation of body and control using `evolution+learning', dividing our review into
synchronous and asynchronous approaches. Asynchronous approaches are synonymous with  the  canonical \textit{steady-state} EA \cite{syswerda1991study}, in which offspring replace one or more member of the population as soon as they are produced. On the other hand, synchronous approaches can be equated to a \textit{generational} model \cite{syswerda1991study} of EA in which the entire population is replaced in one update, selecting from the combined pool of parents and offspring.

Within the class of synchronous approaches, 
Jelasavic \textit{et. al.} \cite{jelisavcic2019lamarckian} tackle the problem of co-optimisation with learning using a modular robotic framework in which body-plans consist of an arrangement of pre-designed modules, and control is realised via a central pattern generator.  A population-based learning algorithm is applied for a fixed number of iterations to improve this population which is in fact stored on the genome. The approach permits both Darwinian and Lamarkian evolution, depending on whether the learned population is written back to the genome. Their results show perhaps unsurprisingly that the Lamarkian approach considerably reduces the time required to learn, and that is particularly important when the available learning budget is small.  Their EA uses non-overlapping generations, i.e. a  new generation of parents is selected from the offspring only via tournament selection, hence this corresponds to a synchronous replacement of the oldest to use the terminology from section \ref{sec:intro}.
Miras \textit{et. al.} \cite{miras2020evolving} follow a similar set up to that described above, using a learning mechanism to tune an inherited controller (without Larmarkianism), finding the learning
influenced the resulting performance and morphological properties of the evolved robots. In this case, the entire population is updated at each generation (i.e. synchronous) via tournament selection from the set of (parents plus offspring), i.e. replacement of the worst.

The ARE framework proposed in \cite{hale2019robot} which aims to evolve robots that can eventually be autonomously manufactured also uses a synchronous framework: here only the body-plan is evolved, with a learning loop  (CMA-ES \cite{hansen2011cma}) applied to each new design, either \textit{tabula-rasa} \cite{le2020sample} or with the initial start point bootstrapped from an stored archive of previously encountered controllers \cite{le2022morpho}. As in \cite{jelisavcic2019lamarckian}, a generational EA is used in which the whole population is replaced by a new population of offspring at each generations. All of the methods just mentioned use a performance based objective function.

Asynchronous methods are less common. Gupta {\em et. al.} \cite{gupta2021embodied} decouple learning and evolution in a distributed asynchronous manner using a tournament-based steady-state evolutionary framework. An EA evolves body-plans with reinforcement learning applied to each new offspring to learn a controller, using a performance based fitness.  After initialisation, tournaments are conducted in groups of 4, with 288 tournaments running in parallel on 288  worker machines.  A single child is produced as a result of each tournament whose training is further distributed over 4 cores. 1152 CPUs are thus required. As soon as a child is trained, it replaces a member of the current population: an ageist replacement mechanism is used, removing the oldest member of the population. Experiments demonstrate that this leads to evidence of a morphological Baldwin effect, demonstrated experimentally by a rapid reduction in the learning times required to achieve a pre-defined level of fitness over multiple generations. The method can hence be classified as (asynchronous, goal-based, oldest).

Finally, although not directly related to evolutionary robotics, Harada and Takadama \cite{harada2020analysis} investigate the dynamics of \textit{semi-synchronous} multi-objective evolutionary framework. Their architecture
generates new solutions whenever evaluations of a \textit{predefined} number of solutions complete, with the level of asynchrony determined by a user-set parameter.
They find that semi-asynchronous approaches with judicious choice of asynchrony can outperform both asynchronous and the synchronous ones, but that the appropriate setting  varies with both target and on the degree of the evolution process.

\section{Methods}

To jointly optimise the design and the controller of a robot, we propose an algorithm structured as a nested optimisation process: an \textit{evolutionary} algorithm optimises the morphology of the robots and a \textit{learning} algorithm optimises the controller for each morphology produced. The algorithm used in the present paper is a modified version of an algorithm called MELAI (morpho-evolution with learning using archive inheritance), first introduced by Le Goff {\em et. al.} \cite{le2022morpho}. We propose a number of modifications to this method which has more degrees of freedom in the evolutionary operators than the original version and a more efficient parallelisation, drawing inspiration from the work of Harada \& Takadama \cite{harada2020analysis} and Gupta \cite{gupta2021embodied}.

\subsection{Evolutionary Algorithm}\label{sec:ea}

The evolutionary algorithm (EA) is structured with two \textit{pools} or populations: the \textit{parents' pool} and the \textit{learning pool} (see figure \ref{fig:gen_dia}). Both pools have the same size. The parents' pool contains the set of robots used to produce offspring, i.e. robots in this pool have been evaluated as a result of undergoing learning. 
Each new offspring is added to the learning pool, which therefore contains a set of robots  either waiting to learn or undergoing a learning process. When a robot finishes its learning, it is added into the parents' pool and removed from the learning pool. To keep the  size of the parents' pool constant,  a robot must be removed from this pool to make space for the new child. This is denoted the \textit{removal} step: robots are removed based on their \textit{age} or their \textit{task performance}. Following this, in order to also keep the learning pool size constant, new robots are generated by mating two robots from the parents' pool. To choose the parents, a tournament is conducted with four randomly selected robots, using either a goal-based objective function, i.e. the task performance or a novelty score rewarding novelty with respect to the robot's design. The genomes of two best robots from the tournament generate a new genome by crossover and then mutation. The new genome is decoded to produce a new robot and then it is added to the learning pool. This is the denoted the \textit{mating} step. 

The \textit{removal} and \textit{mating} steps are applied when a fixed number of robots $N$ have been added to the parents' pool and removed from the learning pool. If this number $N$ is equal to the size of the parent population $P$, i.e. all $P$ robots have completed learning such that there are now $2P$ robots with assigned fitness (parents + offspring), then the algorithm is fully synchronous ($N=P$ in figure~\ref{fig:gen_dia}). On the other hand, if both steps are applied after a single robot completes its learning, then the algorithm is fully asynchronous ($N=1$ in figure~\ref{fig:gen_dia}). In this paper, we do not consider intermediate states of synchronicity as proposed in the work of Harada \& Takadama \cite{harada2020analysis} but leave that for future work.

We experiment with three possible variations of this algorithm that use different instantiations of the key steps: 
(1) synchronous versus asynchronous update, (2) removal of the oldest individuals versus removal of the individuals with the worst task performance of the parents pool, and (3) a goal-based versus novelty-based objective function. 
The combination of synchronous update, removal of the oldest and goal-based objective corresponds to the canonical generational EA model where the whole population of parents are replaced after the children finished being evaluated, i.e. after learning is completed in this case. This variant is used as our baseline as it has been studied in a previous work \cite{le2022morpho}. 


\subsection{Body-plans}

\begin{figure}[h]
    \centering
    \includegraphics[width=0.4\textwidth]{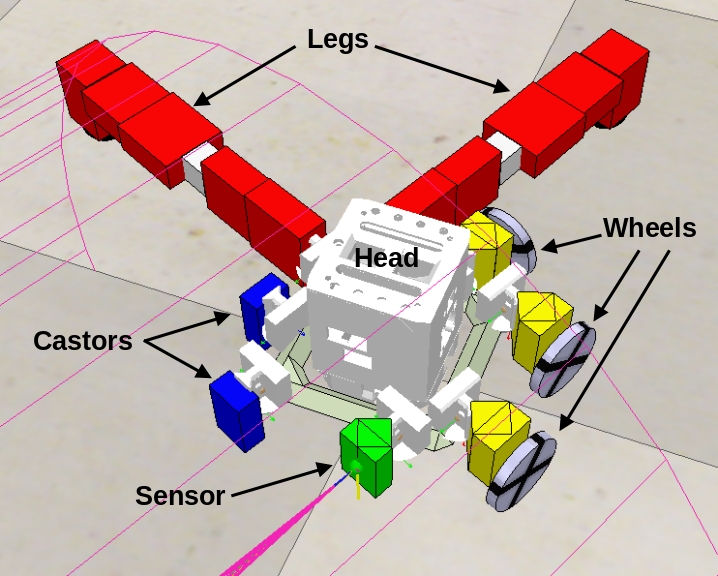}
    \caption{An example of evolved body-plan with a the components labelled. The components are attached on the surface of chassis with the white connectors and the head is attached in the centre of the chassis.}
    \label{fig:rob_example}
\end{figure}

All the body-plans' robot are composed of two parts: a chassis and a set of components (actuators and sensors) that can be attached to the chassis (see figure~\ref{fig:rob_example}). The shape of chassis is evolved but it has  always an embedded computer (dubbed the head) which acts as the brain module  containing the controller software\footnote{This computer (a Raspberry Pi) is only used in experiments with physical robots but it is also included in the simulation to ensure better transfer of robots from simulation to reality.} There are 4 component types: wheels, legs, proximity sensors and castor-balls.

Each body-plan is encoded indirectly by a compositional pattern-producing network (CPPN) \cite{stanley2007compositional}.  When decoding, the coordinates of a 3d matrix are used to query the  CPPN which returns values indicating whether material should be placed at a location. After all positions are queried, a  repair function ensures that the chassis is printable, e.g. removing disconnected plastic and/or overhangs\footnote{This function is strictly only  required if the intention is to build a physical robot}. Additional outputs indicate whether (and where) components are attached to the chassis. An additional viability test discards any robots that do not have any sensors or actuators as they would be unable to move or sense the world.


When using the novelty-based objective function, the novelty score is computed using the method introduced by Lehman {\em et. al.}\cite{lehman2011abandoning} that compares a descriptor defining characteristics of a solution to its $k$ nearest-neighbours. In this paper, we use a novelty score to reward novel body-plans in terms of their shape and attached components. 

To compute this score, a \textit{morphological descriptor} describing the robot's body-plan and a measure of defining distance between two  descriptors is required. The morphological descriptor is a 3d matrix in which the cells represents coordinates. Each cell has a value between 0 and 4: 0 if no component is present and 1 to 4 representing the different types of component. The distance metric used aims to measure the difference in the component set of two body-plans. First, all the components which are different are listed. For each different component located at the same position in both body-plans, the arbitrary value of 11\footnote{11 is the dimension of the matrix} is added to the distance value. Then component at different locations are paired up to the closest one in the other body-plan. The Manhattan distance between both components is added to the distance value. 

The following learning algorithm optimises the controller of the resulting body-plan.

\subsection{Learning algorithm}

Controllers are specified by an Elman controller, i.e. a recurrent neural network \cite{elman1990finding}, with 6 hidden neurons and 6 context neurons.
A modified version of Covariance Matrix Adaptation Evolutionary strategy (CMA-ES) \cite{hansen2016cma} called Novelty-driven with increasing population evolutionary strategy (NIP-ES)  proposed by \cite{le2020sample,le2022morpho} is used to learn the weights of this controller.

NIP-ES is based on the increasing population variation of CMA-ES \cite{auger2005restart} which detects when the algorithm has possibly fallen in a local optimum and then increases the population size to expand exploration. The starting population has 10 controllers: this population is doubled after each restart of the algorithm.  A restart is triggered when the \textit{best fitness} and \textit{behavioural descriptor } metrics stagnate. The stagnation of the \textit{best fitness} metric is triggered when its variance over 20 iterations is less than 0.05. The \textit{behavioural descriptor} stagnation is triggered when its variance across the population is less than 0.05.

NIP-ES combines the behavioural novelty score ($N$) and the task performance ($T$) in a weighted sum as the objective function (see equation \ref{eq:nipesobj}). The factor $n$  defines the novelty ratio. Following \cite{le2022morpho}, we initialise this ratio at 1, i.e. pure novelty search, and it is decreased at 0.05 intervals after each iteration. The ratio is reset to 1 after the algorithm is restarted. 

\begin{equation}\label{eq:nipesobj}
    F(x) = n*N + (n-1)*T
\end{equation}

The novelty score used in NIP-ES is different from the one used in the evolutionary process. It rewards novel behaviour using a \textit{behavioural descriptor}. This descriptor is an 8x8  matrix with binary values. The value of 1 indicates the areas the robot has visited during simulation (see section~\ref{sec:task}). The distance between two descriptors is the squared norm. 

The learning process has two termination conditions: (1) after exhausting a budget of 200 evaluations\footnote{The learning process can exceed this budget only if it reached the maximum number of evaluations in the middle of an iteration. In this case, it will finish the current iteration and then terminates.}, (2) if the robot did not move from its initial position in 50 consecutive evaluations\footnote{the simulation is stopped after 10 seconds if the robot does not move}. 

A \textit{controller archive}, which stores the best controllers according to the combination of the components, is used to speed up learning, following the idea introduced by \cite{le2022morpho}.  Each robot can be described by a descriptor that identifies the number of wheels, legs and sensors attached to the robot which enables any robot to be mapped to a location with a multi-dimensional archive. The \textit{controller archive} is updated after each learning process terminates. If the learning process produced a new better controller than the one available in the archive is replaced. The learning process for a new robot starts with a controller from the archive if there is one available, otherwise is initialised at random. 

\subsection{Parallelisation}
The large number of evaluations required to reach convergence is mainly due to the nature of having a nested optimisation process. Therefore parallelisation can be used to reduce the convergence time.  However, the parallelisation process for this algorithm is complicated by the fact that the learning algorithm can dynamically increase its population size in a manner that cannot be predicted beforehand, so the most efficient way to distribute evaluations is not known \textit{a priori}. We propose the following:

Consider a learning pool of \textit{P} robots just freshly added. All the robots start learning with a population of \textit{C} of controllers. Hence, there are \textit{PxC} pairs of robots and controllers to evaluate in the initial learning step before  moving to the next iteration of learning. All  pairs cannot be evaluated simultaneously if the number of \textit{M} cores available is less than \textit{PxC}. Hence this is mitigated by distributing a portion of controllers of each of the \textit{P} robots to each of the \textit{M} cores. In the case of  \textit{M} less than \textit{P}, some robots will therefore have to wait before starting learning. 

Also, note that the number of controllers evaluated per learning iteration might differ between robots due to the nature of the increasing population mechanism in NIP-ES described in the previous section\footnote{Note that the total number of evaluations remains constant therefore the number of learning \textit{cycles} is variable}. In this case, the parallelisation process randomly selects a pair to distribute, prioritising those that became available for evaluation first.


In the experiments described, we use a population of 25 robots and an initial population of 10 controllers for learning. This requires 250 evaluations alone just for the initialisation step. which is greater than the number of cores at our disposal. This motivates the study of an asynchronous approach in which the population \textit{P} can be updated as soon as a single robot finishes learning.



\subsection{Exploration Task}

\begin{figure}[h]
    \centering
    \includegraphics[width=0.4\textwidth]{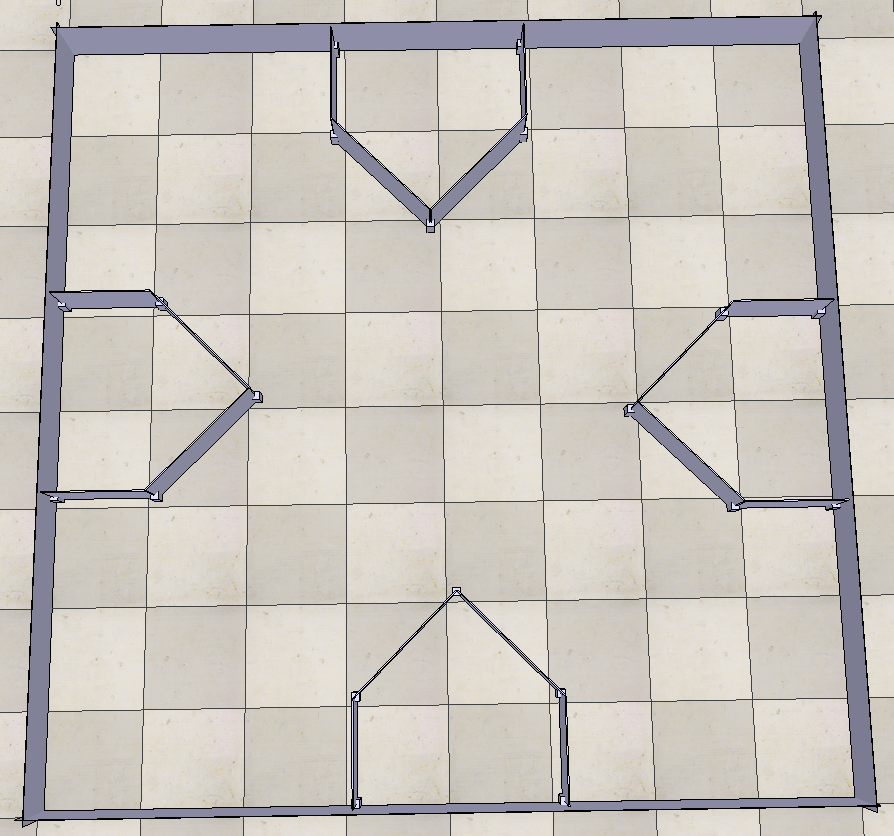}
    \caption{The arena used for the exploration task.}
    \label{fig:arena}
\end{figure}

We conduct a study to investigate the influence of (synchronicity, removal mechanism, and reward) on an exploration task as shown in  figure~\ref{fig:arena}. The environment is divided into 64 tiles. The goal of the task is to maximise the number of tiles visited by a robot within 60 seconds. The goal-based objective function is computed by counting the number of tiles visited divided by 64 (total number of tiles), with the goal of maximising this quantity. Note that there are 16 tiles which are not accessible in the environment, and that the ultimate goal is for the robot to visit as many regions of the environment as possible\footnote{In practice, we observe that 
a robot that visits between 20 and 25 tiles in fact reaches all regions of the environment, where a zone can be considered as quarter of the space}.

The simulator CoppeliaSim's V-REP 3.6.2 is used for the evaluations. 

\section{Results}

For the results shown in this paper, the following convention is followed. Each variant of the algorithm is named with three letters where the first letter indicates if the update step is asynchronous (A) or synchronous (S). The second letter represents whether a robot is removed when it is the oldest (O) or the worst (W) in the generation. The third letter represents whether the objective function is goal-based (G) or novelty-based (N). We evaluate five  different variants of the algorithm (SGO, AGO, SGW, AGW, and ANW) where each variant is evaluated over 15 replicates. 
SGO is considered as the baseline as it follows the architecture from previous publications (e.g. \cite{le2022morpho}). This algorithm is a generational algorithm in which the entire pool of parents is replaced after all the offspring have finished learning. 

\subsection{Task Performance}



\begin{figure}[h]
    \centering
    \includegraphics[width=0.5\textwidth]{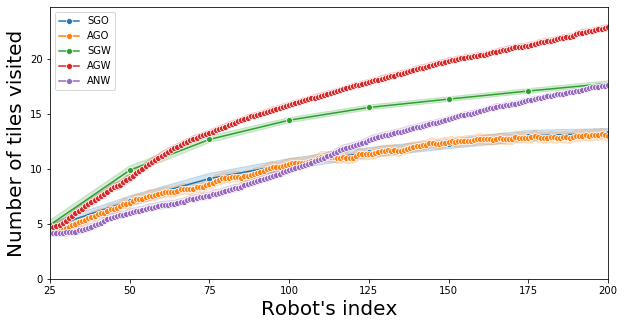}
    \caption{The average number of  tiles visited by robots in the parent pool. The robot's index corresponds to the number of robots evaluated. The dots indicate when the parents' pool was updated.}
    \label{fig:fitness_pool}
\end{figure}

We first compare the variant algorithms in terms of task performance, i.e.their ability to explore the environment. Figure~\ref{fig:fitness_pool} shows the mean task performance over robots present in the pool, ordered
by robot\textit{index}:  this index indicates the number of robots that have been added to the parents' pool. For the asynchronous methods (AGO, AGW, ANW),  one robot is added at each update and therefore there is one data point for each index. In contrast, in the synchronous methods (SGW, SGO), the pool is only updated once $P$ new robots are produced, i.e. every 25 robots. In this case, there is a data point every 25 robots.

The best parameterisation of the is the combination of asynchronous, removal of the worst and goal-based objective (AGW), showing a rapid increase after evaluation of the initial population.  The worst performing combinations are AGO and SGO, i.e. variants that use goal-based rewards and replacement of the oldest.  The  variant ANW that uses a novelty-based objective progresses slowly, but eventually catches SGO, ending in second position. The curve is still rising at the end point of the experiment, suggesting that the novelty-based approach might reach better fitness if run longer although is unlikely to reach the performance of AGW.  Note however that robotics experiments are often limited by a specific evaluation budget given the time taken for evaluations, hence the `any-time' performance of an algorithm is often an important factor in choice of algorithm.
The "W" variants that replace the worst member(s) of the population all clearly outperform the baseline, with AGW and SGW showing particularly fast increases in performance in the early stages of evolution.
When comparing SGO and AGO, the move to an  asynchronous update does not improve the performance. In contrast, when comparing SGW and AGW,  a significant increase in performance is gained by switching to an asynchronous update\footnote{Statistical test applied to the performance values at the end point: Mann-Whitney U test p-value $2*10^5$ and critical value 9 (< 64)}. 
In summary, replacement of the worst is always beneficial regardless of synchronicity. The
 novelty-based objective (ANW) leads to better performance than the baseline (SGO), while a goal-based objective leads to better performances when associated with asynchronous updates and removal of the worst. 

\subsection{Morphological variance}\label{sec:morph_var}

\begin{figure}[h]
    \centering
    \includegraphics[width=0.5\textwidth]{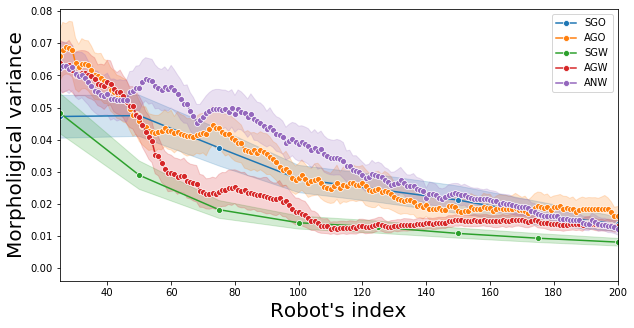}
    \caption{The morphological variance over the robot's index. The robot's index corresponds to the number of robots evaluated. The morphological variance is computed based on a descriptor of the body-plan design comprising: the number of wheels, legs, sensors, and the height, width and depth of the chassis. The higher the value, the higher the variance.}
    \label{fig:morph_var_pool}
\end{figure}

Figure~\ref{fig:morph_var_pool} shows the variance of robots' body-plan within a parents' pool over the robot's index. The morphological variance is computed by first summing the normalised numbers of wheels, legs, sensors, and casters and the normalised height, depth and width of the chassis to obtain a unique value characterising the design of the robots. Then, the \textit{morphological variance }is defined as the variance of this metric over the individuals in the parents' pool. 

Figure~\ref{fig:morph_var_pool} shows the diversity change of body-plan designs during the evolutionary process. There is a high variance in the initial population due to the random initialisation process. The AGW and SGW methods undergo a sharp drop in variance following initialisation and converge quickly to  a plateau. This results from the elitist nature of the update mechanism which replaces the worst robot(s) in the pool, thereby decreasing diversity. The methods AGO and SGO undergo a slower decrease in diversity: as noted by Gupta {\em et. al.}, using an ageist replacement policy that removes the oldest members helps maintain diversity.  Finally, the variant ANW that uses a novelty score as an objective function maintains  morphological diversity for a longer period that any of the other methods, but eventually converges to a similar morphological variance as the other approaches at the end point.

As expected, the methods that remove the oldest robot and that use a novelty-based goal as an objective are able to better protect the diversity of body-plan design over the optimisation process. There is no observable difference in diversity between the asynchronous and the synchronous update methods.  Although all the approaches unsurprisingly lead to a loss of diversity over time, the differing combinations of (synchronicity, removal, objective) exhibit different evolutionary dynamics in terms of the diversity metric. Specifically, the strategies that remove the oldest robot and/or use novelty-based objective features enable more exploration of new designs before converging. 



\subsection{Analysis of the highest performing robots}
As the motivation behind our study is ultimately to produce high-performing robots, in this section we  provide an analysis of on the 20 top robots produced by each replicate for each variant during the entire evolutionary process (i.e. regardless of the generation at which the robot was found).

\begin{figure}[h]
    \centering
    \includegraphics[width=0.4\textwidth]{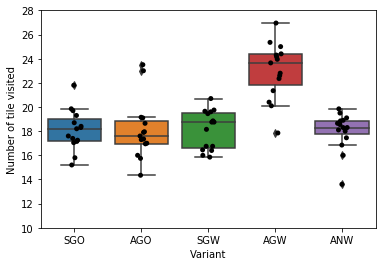} \\
    \includegraphics[width=0.4\textwidth]{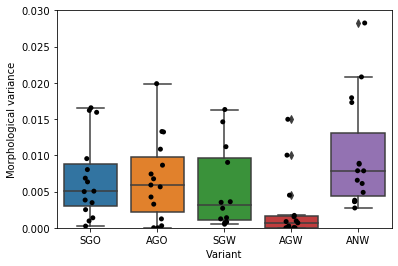} \\
    \includegraphics[width=0.4\textwidth]{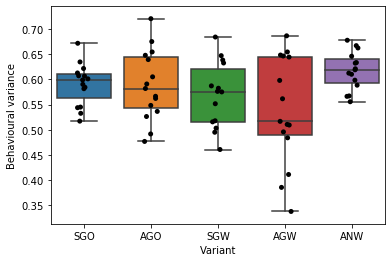} 
    \caption{The morphological and behavioural variance for each variant of the 20 best robots (selected according to their task performance). 
    The morphological variance is computed based on a descriptor of the body-plan design comprising: the number of wheels, legs, sensors, and the height, width and depth of the chassis.  The behavioural variance is computed based on the trajectory of the robots of their best controller.}
    \label{fig:morph_beh_var}
\end{figure}

Figure~\ref{fig:morph_beh_var} shows the mean performance (i.e. the number of tiles visited), the morphological and behavioural variance of the 20 best robots for each replicate. The \textit{morphological} variance is computed in the same manner as previously described in section~\ref{sec:morph_var}. 

The \textit{behavioural} variance of the 20 best robots corresponds to the average of the distances of the trajectories between each robot. The distance is computed in the following manner. For any pair of robots, the distance between their trajectories is defined as the average of the distances between each of $t=180$ points (represented as 2d coordinates) in the environment  visited during an evaluation. This is computed for all possible pairs of trajectories and then the mean is taken, hence this quantity represents the  average variation between trajectories.

The algorithm variant AGW (asynchronous update, goal-based objective, and removal of the worst) tends to provides the best set of high-performing robots (task performance) across the the majority of replicates. However, it is the  parameterisation that delivers the lowest morphological variance. The other combinations have similar distribution in terms  of task performance and morphological variance. The variant ANW using the novelty-based objective has a significantly higher morphological variance than its counterpart using goal-based objective (AGW) (see second plot figure~\ref{fig:morph_beh_var})\footnote{Statistical test Mann-Witney U test: p-value of 0.02 and critical value 61 (< 64)} The baseline SGO  achieves second place in terms of morphological variance. However, no parameterisation provides a consistent guarantee of both high diversity and high performing robots.

\begin{figure}[h]
    \centering
    \includegraphics[width=0.5\textwidth]{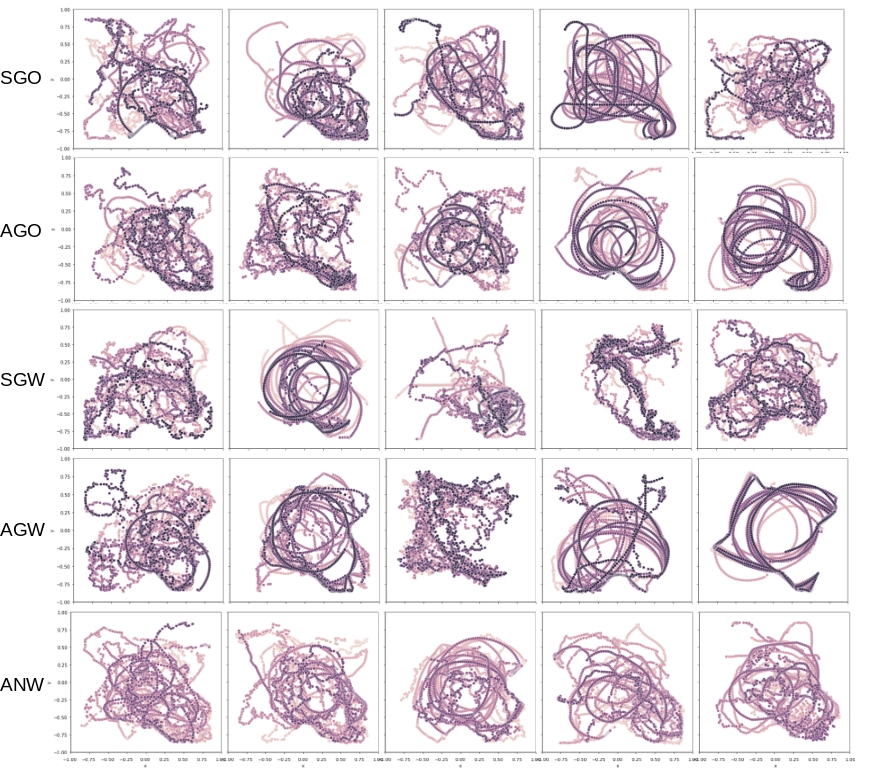} 
    \caption{Trajectories of the 20 top robots for 5 replicates of each variant.}
    \label{fig:traj}
\end{figure}

In term of behavioural variance, there is little difference between the variant algorithms. They all result in a diversity of behaviours (see figure \ref{fig:morph_beh_var,fig:traj}.  However AGW produces some replicates  with very low behavioural diversity (figure \ref{fig:morph_beh_var}). 

\begin{figure}[h]
    \centering
    \includegraphics[width=0.45\textwidth]{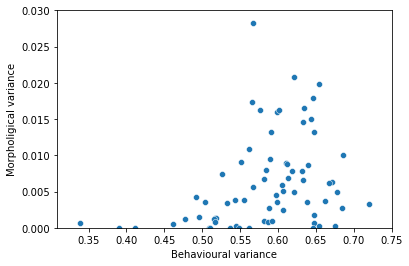} 
    \caption{The figure shows behavioural variance plotted against morphological variance of 5 replicates of the top 20 robots.}
    \label{fig:beh_vs_morph}
\end{figure}

Figure~\ref{fig:traj} provides a visual  demonstration of the behavioural variance that was captured in a single metric in the previous plot for the 20 best robots for 5 replicates for each variant. Each cell represents a single replicate and contains the trajectories for each of 20 robots. Trajectories in the 5th column for AGO, 2nd column for SGW and 5th column for AGW are similar and appear to have low behavioural variance.  Some of the trajectories shown are obviously noisy  --- these were all produced with robots with legs. For instance, the 3rd column of AGW shows only noisy trajectories. In contrast, robots with wheels tend to produce smoother trajectories.  Cells with a mixture of noisy and smooth trajectories  contain morphologically diverse robots, i.e. a high morphological variance leads to high behavioural variance. Moreover, figure~\ref{fig:beh_vs_morph} which plots behavioural variance against morphological variance for replicates of the best 20 robots shows that even with low morphological variance, the algorithm is able to produce high performing robots with diverse behaviours (high behavioural variance).   

Moreover, as expected, the variant using the novelty-based objective (ANW) provides more replicates with diverse trajectories: recall that the novelty metric rewards morphological diversity which in turn lead to behavioural diversity.



\section{Discussion and Conclusion}

The experiments described above compare five variants of an evolutionary algorithm (SGO, AGO, SGW, AGW and ANW) that aim to investigate the influence of three factors on task performance and diversity: synchronicity (S) or asynchronicity (A), removal of the oldest robot (O) or the worst robot (W), and the reward mechanism goal-based (G) or novelty-based (N).
From the experimental results presented in the previous section, the variant AGW (asynchronous updates, goal-based objective, and removal of the worst) outputs the best performing solutions. SGW --- which differs from AGW  only in that it performs a synchronous update of the entire pool --- follows a similar trajectory to AGW  in the early part of the evolutionary process making rapid improvement, but converges to a lower performance value.
The early progress is explained by the elitist replacement mechanism (replace worst) and goal-based objective.
However, for AGW, it comes at the price of lower diversity in the best robots generated.

Using novelty score (ANW) instead of the task performance as an objective does not lead to improved performance compared to AGW. In other literature (e.g. \cite{lehman2011abandoning,mouret2015illuminating}), searching for novelty has been demonstrated to lead to higher overall performance due to the increased exploration of the search space but this is not apparent in our results. However, if examining individual replicates, it can lead to high performing robots in some cases. 
Combining an elitist replacement mechanism with a novelty-based reward function (rather than the more elitist approach of rewarding performance)
can lead to a balance diversity and performance like shown in the quality-diversity literature \cite{lehman2011abandoning,mouret2015illuminating}, however, this is not consistently shown in our experiments.   


Interestingly, the asynchronous update mechanism leads to better performance than synchronous updates when associated with removal of the worst (SGW vs AGW)  but to similar performances when associated with removal of the oldest (SGO vs AGO).  Recall that with an asynchronous updates framework,  robots from the parents' pool are replaced one by one, while the synchronous update results in 25 robots being replaced at once. With an asynchronous update, in the case of elitist replacement, i.e. removal of the worst, a design with relatively low fitness compared to others in the pool is able to remain in the pool over multiple updates. In contrast, it would likely be replaced in the synchronous update. Thus it can remain available for selection for mating over multiple generations. This acts similarly to the \textit{innovation protection} mechanism proposed by Cheney {\em et. al.} \cite{cheney2018scalable}. They noted that evolutionary changes to morphology can adversely impact control, leading to poor behavioural performance. They introduced a technique that leads to ‘morphological innovation protection’ that temporarily reduces selection pressure on recently morphologically changed individuals to enable evolution or learning to have time to re-adapt control. In the case of non-elitist replacement, i.e. removal of the oldest, the benefit of asynchronous updates is not apparent as the replacement mechanism acts as a stronger innovation protection.


It is also interesting to compare SGO (synchronous update, goal-based objective, and removal of the oldest) and AGW (synchronous update, goal-based objective, and removal of the worst) as the first corresponds to the canonical generational EA model and the second to the steady-state EA model. Our results show that AGW leads to better performance than SGO as the first is more elitist (replacing the worst robots from the parents pool). On the other hand, SGO replaces the oldest robots and therefore less high performing robots are able to remain in the pool. 

SGO maintains a higher morphological diversity in the designs than AGW during the evolutionary process (figure~\ref{fig:morph_var_pool}) but when comparing the best solutions found from both method, the morphological diversity is similar (figure~\ref{fig:morph_beh_var}). Hence, the steady-state (asynchronous) EA  (AGW) with an elitist replacement mechanism leads to better performance than a generational (synchronous) EA (SGO).

To summarise:
\begin{itemize}
    \item \textit{Removal of the worst} (W) leads to higher performing robots than \textit{removal of the oldest} (O) combined with any of the other variations of synchronicity or reward.
    \item \textit{Novelty-based objective} associated with \textit{removal of the worst} (ANW) does not outperform its counterpart using goal-based objective (AGW), i.e. searching for novelty does not appear to increase exploration of the search-space.
    \item With the parallelisation mechanism proposed, \textit{asynchronous updates} result in better performance than \textit{synchronous updates} when associated with \textit{removal of the worst} and \textit{goal-based objective}, i.e. AGW vs SGW.
\end{itemize}

Although the majority of the discussion has focused on performance and diversity, it also worth noting that the asynchronous method (AGW) leads to rapid progression in terms of fitness, hence provides better \textit{any-time} performance. This is particularly important in robotics where evaluations can be very time-consuming (and particularly if the intention is to eventually conduct evaluations on real robots).

The scope these conclusions is of course  limited to the joint optimisation of robot's design and controllers in the particular morphological space we describe here, the task and the  learning algorithm chosen.
 In future work, we intend to extend this investigation to a broader range of tasks such as targeted locomotion or locomotion on rough terrains in order to test the generality of our conclusions. Finally, it would also be interesting to consider approaches to parallelisation that lie between the fully asynchronous/synchronous frameworks evaluated here. For example, the semi-synchronous approach proposed by Harada and Takadama \cite{harada2020analysis}  would be worth evaluating, even though this was shown to require careful tuning of the synchronicity parameter.

 \clearpage
\bibliographystyle{acm}
\bibliography{references}

\end{document}